\documentclass[10pt,twocolumn,letterpaper]{article}

\usepackage{cvpr}
\usepackage{times}
\usepackage{epsfig}
\usepackage{graphicx}
\usepackage{amsmath}
\usepackage{amssymb}
\usepackage[frozencache,cachedir=.]{minted}

\usepackage[breaklinks=true,bookmarks=false]{hyperref}

\cvprfinalcopy 

\def\cvprPaperID{****} 

\begin{document}

\title{Pose Trainer: Correcting Exercise Posture using Pose Estimation}

\author{Steven Chen \thanks{The authors contributed equally to this work.} \hspace{1cm}
Richard R. Yang $^*$ \\
Department of Computer Science, Stanford University
}

\maketitle
\begin{abstract}
Fitness exercises are very beneficial to personal health and fitness; however, they can also be ineffective and potentially dangerous if performed incorrectly by the user. Exercise mistakes are made when the user does not use the proper form, or pose. In our work, we introduce Pose Trainer, an application that detects the user's exercise pose and provides personalized, detailed recommendations on how the user can improve their form. Pose Trainer uses the state of the art in pose estimation to detect a user's pose, then evaluates the vector geometry of the pose through an exercise to provide useful feedback. We record a dataset of over 100 exercise videos of correct and incorrect form, based on personal training guidelines, and build geometric-heuristic and machine learning algorithms for evaluation. Pose Trainer works on four common exercises and supports any Windows or Linux computer with a GPU.
\end{abstract}

\section{Introduction}

Exercises such as squats, deadlifts, and shoulder presses are beneficial to health and fitness, but they can also be very dangerous if performed incorrectly.  The heavy weights involved in these workouts can cause severe injuries to the muscles or ligaments. Many people work out and perform these exercises regularly but do not maintain the proper form (pose). This could be due to a lack of formal training through classes or a personal trainer, or could also be due to muscle fatigue or using too much weight. For our course project, we seek to aid people in performing the correct posture for exercises by building Pose Trainer, a software application that detects the user's exercise pose and provides useful feedback on the user's form, using a combination of the latest advances in pose estimation and machine learning. Our goal for Pose Trainer is to help prevent injuries and improve the quality of people's workouts with just a computer and a webcam.

The first step of Pose Trainer uses human pose estimation, a difficult but highly applicable domain of computer vision. Given visual data, which could be an RGB image and/or a depth map, a trained model predicts a person's joints as a list of skeletal keypoints. Pose estimation is critical for problems involving human detection and activity recognition, and can also aid in solving complex problems involving human movement and posture. We use a state-of-the-art pose estimation deep neural network, OpenPose~\cite{realtimemultiperson}, within Pose Trainer for inference.

The second part of our application involves detecting the quality of a user's predicted pose for a given exercise. We approach this using heuristic-based and machine learning models, using the poses and instruction of personal trainers and other qualified professionals as the ground truth for proper form.

Our full application consists of the previously described two main components, combined into an end-to-end application that can take a video of an exercise and provide useful exercise form feedback to the user. For the scope of this project, we present a Pose Trainer desktop application: future work involves moving Pose Trainer to mobile devices to create a true on-the-go personal trainer.

\section{Problem Statement}

For the pose estimation component, we utilize a pre-trained real-time system, called OpenPose~\cite{realtimemultiperson}, that can detect human body keypoints in videos. (More detail on OpenPose and our reasoning for choosing this system can be found in Technical Approach.) This model is functional out of the box, and thus is very simple to install for users of our application. Using OpenPose allows us to take advantage of the state of the art in pose estimation algorithms for our task, and lets us focus on the actual evaluation of exercise posture.

For the posture evaluation (pose training) component, we have recorded videos of ourselves performing exercises. Our videos include our best effort to correctly perform the exercise, as well as intentionally incorrect examples.

The evaluation of our posture identifier is dependent on the performance of the pose estimator. We work under the assumption that the pose estimator is accurate a majority of the time, with small measurement deviances due to noise, which we correct for. We evaluate our posture identifier in different ways depending on the algorithm: for heuristic algorithms, we feed in all videos for evaluation, while for machine learning algorithms, we evaluate by splitting our video dataset into train and test sets, and report results on the test set.

\section{Related Work}

We identify several state-of-the-art methods for pose estimation that accurately estimate human poses under a variety of sensor configurations, shots, and counts of individuals per shot. Toshev et al.~\cite{deeppose} were the first to use a deep neural network to improve pose detection, finding the location of each body joint using regression on CNN features. Newell et al.~\cite{stackedhourglass} introduce a stacked hourglass neural network architecture that uses repeated bottom-up and top-down processing to achieve accurate single pose predictions. Wei et al.~\cite{cpm} propose a different architecture, using multiple convolutional networks to refine joint estimates over sequential passes. Instead of RGB camera data, Shotton et al.~\cite{singledepth} use single depth maps captured by the Microsoft Kinect to predict 3D positions of joints through an object recognition approach. Bogo et al.~\cite{smpl} estimate 3D pose, as well as 3D mesh shape, using just single RGB images.

A significant area of research has also focused on detecting the poses of multiple people in one shot. Papandreou et al.~\cite{accuratemultiperson} detect multiple poses through a two-stage process, first identifying possible bounding boxes for people, then detecting pose keypoints in each bounding box. In contrast, Cao et al.~\cite{realtimemultiperson} use Part Affinity Fields to estimate poses of multiple people in a scene in real time without the need to identify individual persons first. Cao et al.~\cite{realtimemultiperson} have open-sourced their work as a project called OpenPose, which we utilize for Pose Trainer.

Pose estimation allows us to analyze the static posture of humans, which will provide valuable information regarding posture correctness. Zell et al.~\cite{physicalanalysis} use an interesting approach for analysis of physical movements, where the body is represented as a mass-spring system and used to find the forces and torques that travel through the joints of the body. We have found that, by using exercise specifications and feedback from professionals, we can take a simpler approach to physical analysis, analyzing the angles and distances between joint keypoints to provide important feedback to users without needing a full physical simulation.

\section{Technical Approach}

\subsection{Pipeline Overview}

\begin{figure}[t]
    \centering
    \includegraphics[width=0.99\linewidth]{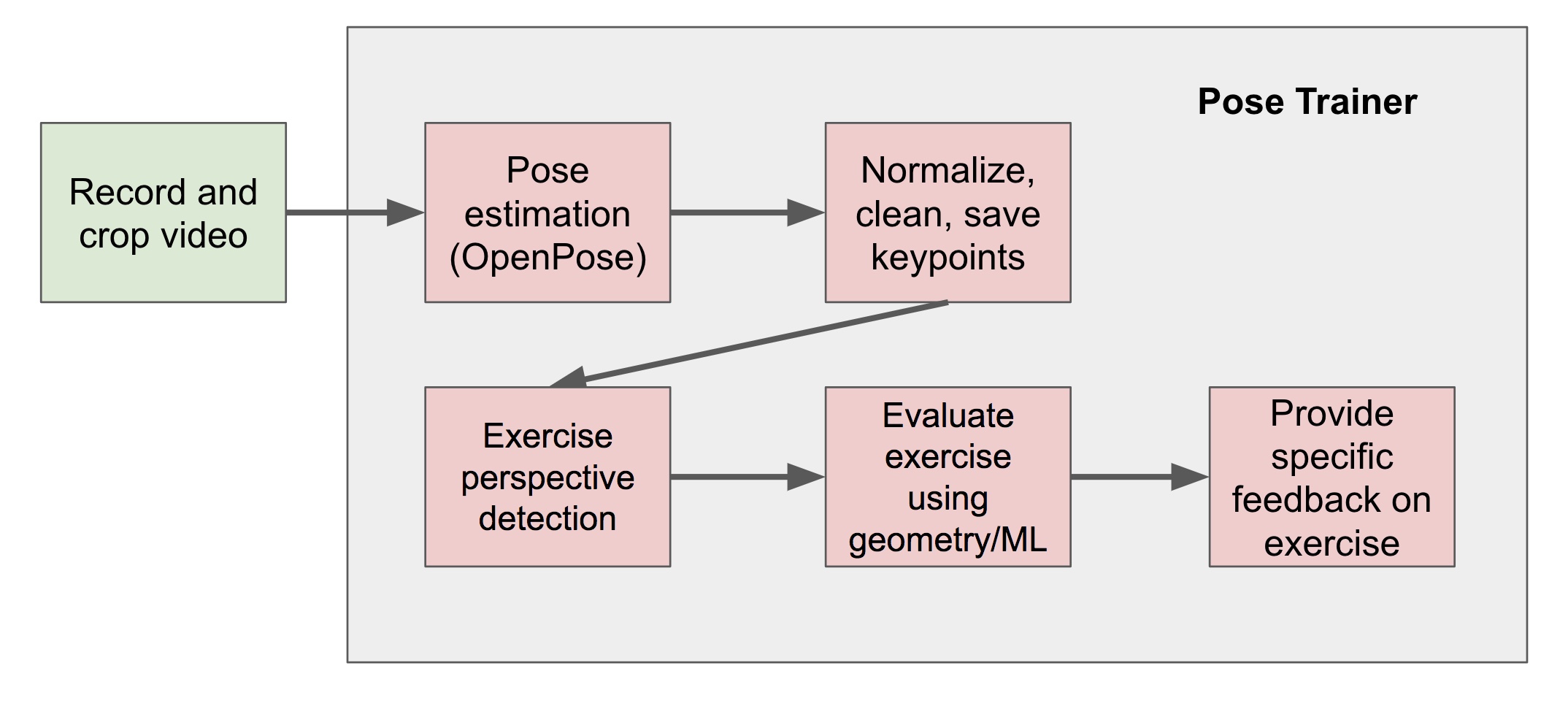}

    \centering
    \caption{Pose Trainer system pipeline, as described in Technical Approach.}
    \label{fig:flowchart}
\end{figure}

We now describe the Pose Trainer application from a technical perspective as a pipeline system, consisting of multiple system stages (see Figure \ref{fig:flowchart}). Pose training starts from the user recording a video of an exercise, and ends with the Pose Trainer application providing specific feedback on the exercise form to the user.

\subsection{Video Recording}

First, the user records a video of themselves performing a selected exercise. The video is recorded from a particular perspective (facing the camera, side to the camera, etc) that allows the exercise to be seen. However, there are no requirements on camera type or distance from camera: the user only needs to make sure that their body is visible. Then, the user trims the video such that it includes only the frames of the exercise.

We leave video recording and cropping to the user, so that they can use whatever video recording and editing system they are most comfortable with. All modern computers and smartphones include easy-to-use video recording and cropping software, and since Pose Trainer supports all common video formats, any software can be used.

\subsection{Pose Estimator}

For pose estimation, we use deep convolutional neural networks (CNNs) to label RGB images. After experimentation with multiple state-of-the-art pose estimators, we choose to use the pre-trained model, OpenPose, for pose detection. OpenPose introduces a novel approach to pose estimation using part affinity fields, which are vectors that encode the position and orientation of limbs. The model is composed of a multi-stage CNN with two branches, one to learn the confidence mapping of a keypoint on an image, and the other to learn the part affinity fields ~\cite{realtimemultiperson}. OpenPose is both accurate and efficient, while also scalable to multiple people without scaling up the run-time.

Another important factor in our choice of OpenPose is its ease of installation and use for our end-users. Most pose estimators are currently released as Tensorflow or Caffe source code, containing the model architecture, and require challenging user installations and downloading of weights to be usable. In addition, GPU libraries such as CUDA and CuDNN need to be installed for these estimators to run properly. This is far beyond the reach of most average computer users. In contrast, OpenPose is released in Windows and Linux executable format, which means that no installation or knowledge of programming is required at all to use it. In addition, no external installation of GPU libraries are required: OpenPose will automatically use the NVIDIA GPU inside the computer, if present. This means that in order to use our application, Pose Trainer, a user only needs to download OpenPose and install Anaconda to get up and running.

OpenPose output consists of lists containing the coordinate predictions of all keypoint locations, and their corresponding prediction confidence. We consider the predictions of 18 keypoints of the pose, which include the nose, neck, shoulders, elbows, wrists, hips, knees, and ankles.

\subsection{Keypoint Normalization}

After running pose estimation, we write our own parser to use the raw keypoint output. First, we read the list of keypoint predictions for every video, and segment the list into Part objects that store the $x$, $y$, and confidence of each keypoint. Additionally, we track if a keypoint is obscured in the image frame (prediction confidence of 0). For every frame in the video, we compose each Part (joint keypoint) to construct a Pose object that represents the full skeletal estimation of a person. For the full video, we compose a PoseSequence object that chains the Pose estimations from each video frame.

We note the need to generalize our application to account for users with different body length measurements, distance from the camera, as well as other relative factors. To account for these differences, we normalize the pose based on the torso's length in pixels. The torso length is calculated by the average of the distance from the neck keypoint to the right and left hip keypoints. This normalization works extremely well: we observe that the torso length stays very constant through all the frames of input videos. Distances are thus represented as ratios of torso length: for instance, an upper arm length of 0.6 means that the upper arm is 0.6 the length of the torso.

\subsection{Perspective Detection}

For certain exercises, we resolve the ambiguity in camera perspective. For example, the bicep curl exercise is recorded from the side of the body, and could be performed with either the left or right arm. We identify which arm is performing the exercise in the video by measuring which keypoints are most visible (left or right side keypoints) throughout all frames of the exercise. This accurately detects perspective all of our input videos.

\subsection{Geometry Evaluation}

Next, we compute body vectors from keypoints of interest, and use personal training guidelines and our own recorded videos to design geometric heuristics, evaluating on the body vectors.

\subsubsection{Example: Bicep Curl}

For instance, in a bicep curl, we identify two heuristics of interest. First, the upper arm should be kept steady and not move significantly. We quantify this by the angle between the upper arm vector and the torso vector. If the upper arm is held steady, then it should be parallel to the torso with minor variations for the entire video. Significant movement of the upper arm will result in a large change in the angle between the two vectors. 

Second, a proper, complete curl requires the weight to be brought up until the bicep is fully contracted, beyond the midway point (90$^{\circ}$ between upper arm and forearm) that is commonly stopped at. This improper form is typically a result of the user using weights that are too heavy, which our application will point out. We quantify this problem by the minimum angle achieved between the upper arm and forearm. In the start position with the weight down, the angle should be nearly 180$^{\circ}$. As the weight is lifted, the angle should decrease until when the user stops, and increase again as the weight is brought down. Thus, finding the minimum angle across the video sequence will show how far the user brought up the weight. 

Through our analysis of annotated video data, we find that if the angle between the upper arm vector and torso vector exceeds 35$^{\circ}$, then the user is rotating their upper arm excessively. If the minimum angle between the upper arm vector and forearm vector is not below 70$^{\circ}$, then the user is not curling the weight all the way up. Using the quantified measures and thresholds, we alert the user specifically which thresholds they exceed, and offer suggestions to improve their form. 

Details on our geometric evaluations of the other exercises are provided in the Results section.

\subsection{Machine Learning Evaluation}

Our second approach to evaluate the exercise posture given normalized keypoints uses a more data-driven, machine learning approach. Since the recorded videos can be of arbitrary length, this results in a different keypoint vector length for each example. Different feature vector lengths present a challenge for many machine learning models, so we approach this task using dynamic time warping (\textbf{DTW})~\cite{dtw} with a nearest neighbor classifier.

DTW is a metric used to measure the non-linear similarity between two time series. When two similar sequences are phase-shifted (i.e. shifted in the time dimension), metrics like the Euclidean distance fail because it is a direct comparison of two points at the same time. In DTW, we try to dynamically identify the keypoint in the second sequence that corresponds to a given point in the first.

Given two keypoint sequences $Q \in \mathbb{R}^m$ and $C \in \mathbb{R}^n$, we construct a distance matrix $D\in\mathbb{R}^{m \times n}$, where $D_{i,j}$ is the Euclidean distance between points $q_i$ and $c_j$. We iterate through all points in the matrix using dynamic programming, find the optimal match of points in each sequence, and compute the distance of between the matched points. A drawback with DTW is that it's not robust to noise. When OpenPose generates noisy keypoints, this would affect the performance of DTW. To accommodate for this, we run the a keypoint sequence through a size 5 median filter twice before computing the DTW measures. 

We take an input keypoint sequence and compute the DTW distance of it with all the training sequences. Then, we build a binary nearest neighbor classifier that predicts 'correct' or 'incorrect' form based on the DTW distances.

\section{Results}

We present quantitative and qualitative results of Pose Trainer on four different dumbbell (free motion) exercises: bicep curl, front raise, shoulder shrug, and standing shoulder press. For each exercise, we take both a geometric/heuristic approach, as well as a machine learning approach using dynamic time warping.

\subsection{Bicep Curl}

The single arm bicep curl is a dumbbell (hand-held free weight) exercise that isolates the biceps, the large muscle in the upper arm responsible for bending in and twisting at the elbow. In the bicep curl, a dumbbell is lifted up from a resting, extended position, with rotation around the elbow, while keeping other parts of the body still. This action targets the exercise towards the biceps muscle. Common mistakes when performing a bicep curl include using the shoulder to help swing the weight up, and thus rotating around the shoulder, as well as not lifting the weight fully up.

In our geometric algorithm, we consider the range in angle between the upper arm and torso (measuring if the user rotates the shoulder when lifting), and the minimum angle between the upper arm and forearm (measuring how high the user lifts). If the range between upper arm and torso angle is above 35 degrees, we flag this as too much shoulder rotation. If the minimum angle between upper arm and forearm is above 70 degrees, this is flagged as not curling the weight all the way up.

Our geometric algorithms correctly classify all good form videos correctly. 80\% of bad form exercises are labeled as incorrect, while the remainder are labeled as correct form. Upon visual analysis, these mislabeled examples are on the fringe between proper form and clearly improper form: tweaking our algorithms will, as expected, adjust the decision boundary, and thus our strictness when evaluating user exercise posture.

For our ML algorithm, we split our 16-video bicep dataset into 9 training examples and 7 test examples. The DTW classifier achieves an F1 score of 0.85. The precision, recall, and F1 for all exercises are summarized in table \ref{tbl:dtw_comparison}.

\begin{figure}[t]
    \centering
    \includegraphics[width=0.4\linewidth]{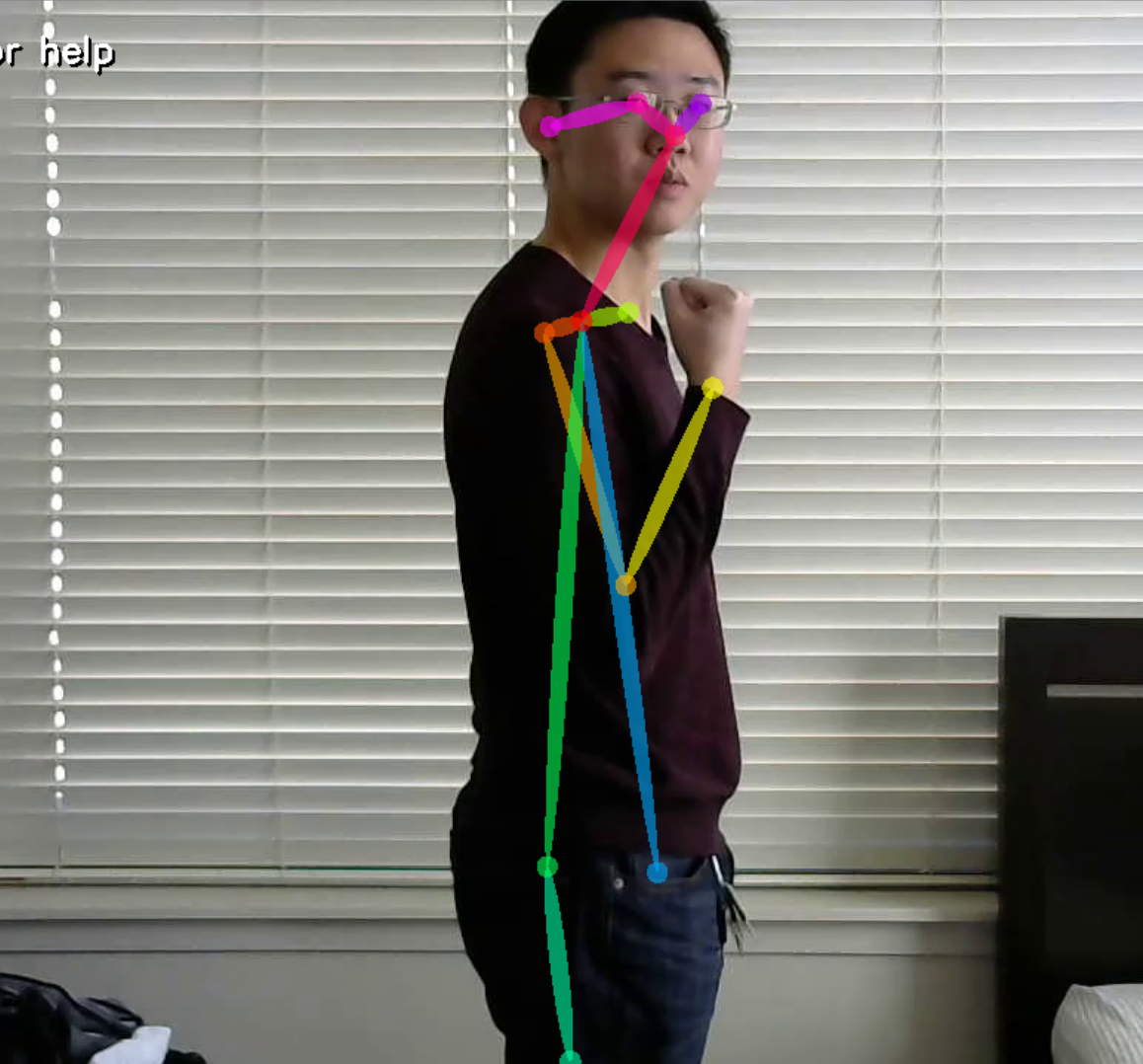}
    \hspace{0.5cm}
    \includegraphics[width=0.38\linewidth]{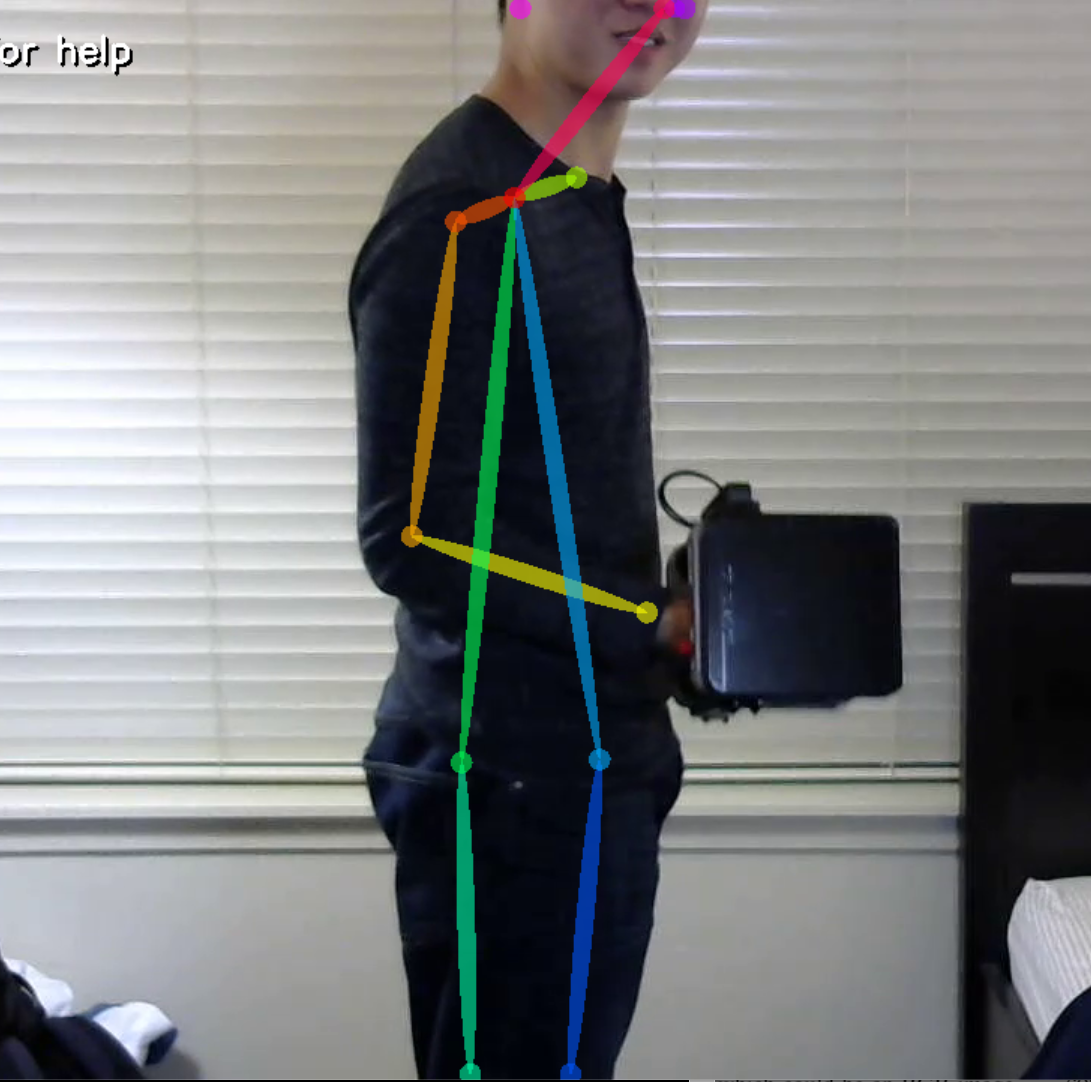}
    
    \includegraphics[width=0.49\linewidth]{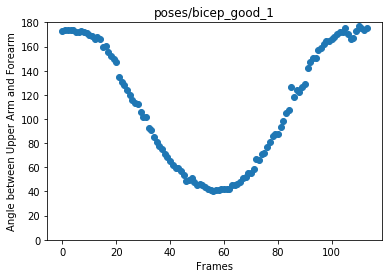}
    \hfill
    \includegraphics[width=0.49\linewidth]{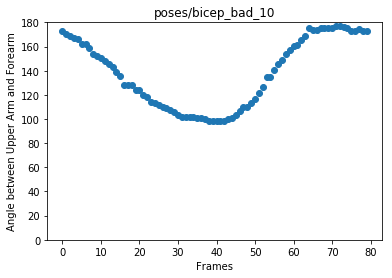}
    \centering
    \caption{In a proper bicep curl, the arm should contract until the forearm is as close to parallel with the torso as possible. Left shows proper midpoint form: right shows improper form, where the exercise has no completely finished. The bottom plots show the angle between the upper arm and the forearm over time: at the midpoint of the exercise, the angle should be small, meaning the arm has fully contracted.}
    \label{fig:bicep_curl_lift_up}
\end{figure}

\begin{figure}[t]
    \centering
    \includegraphics[width=0.4\linewidth]{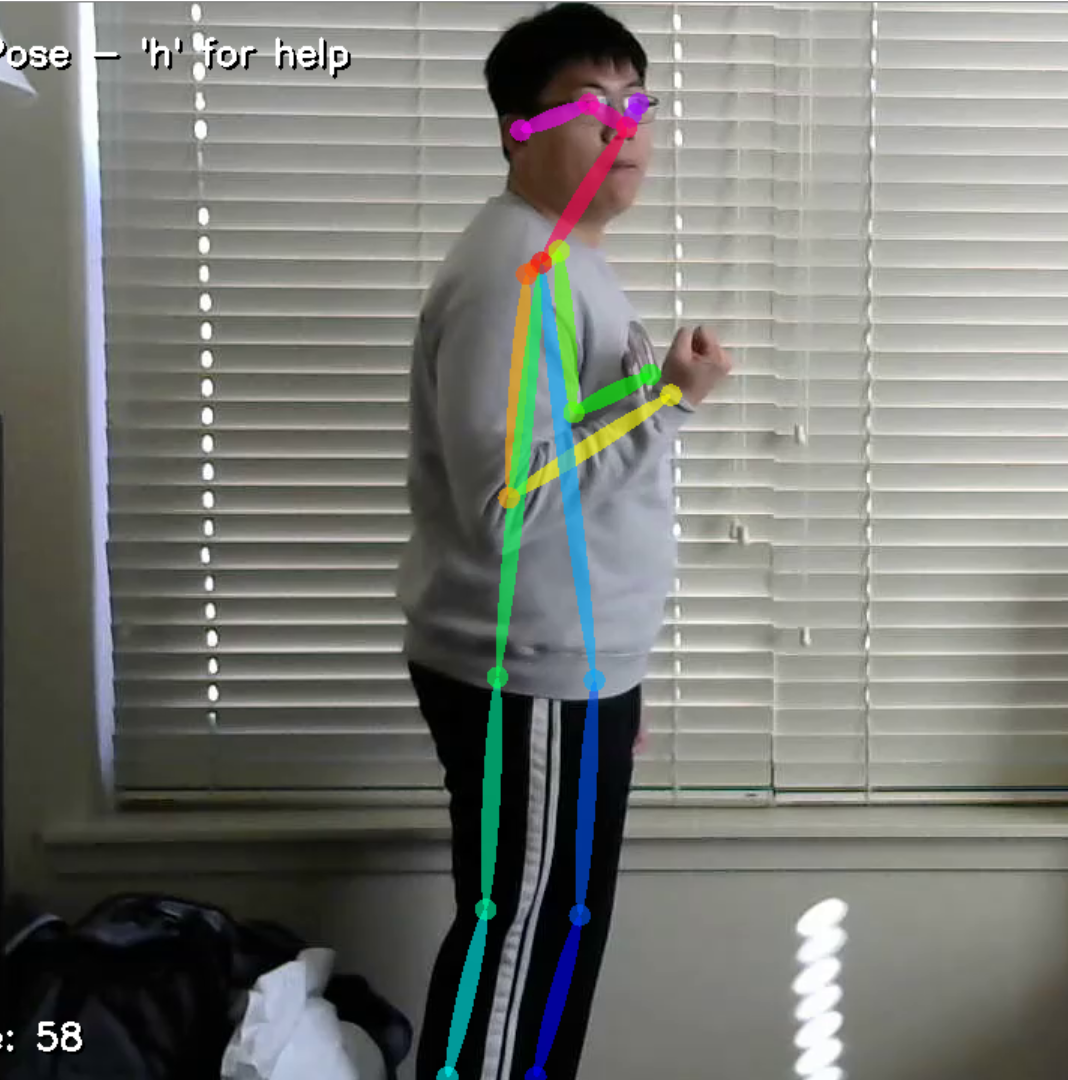}
    \hspace{0.8cm}
    \includegraphics[width=0.31\linewidth]{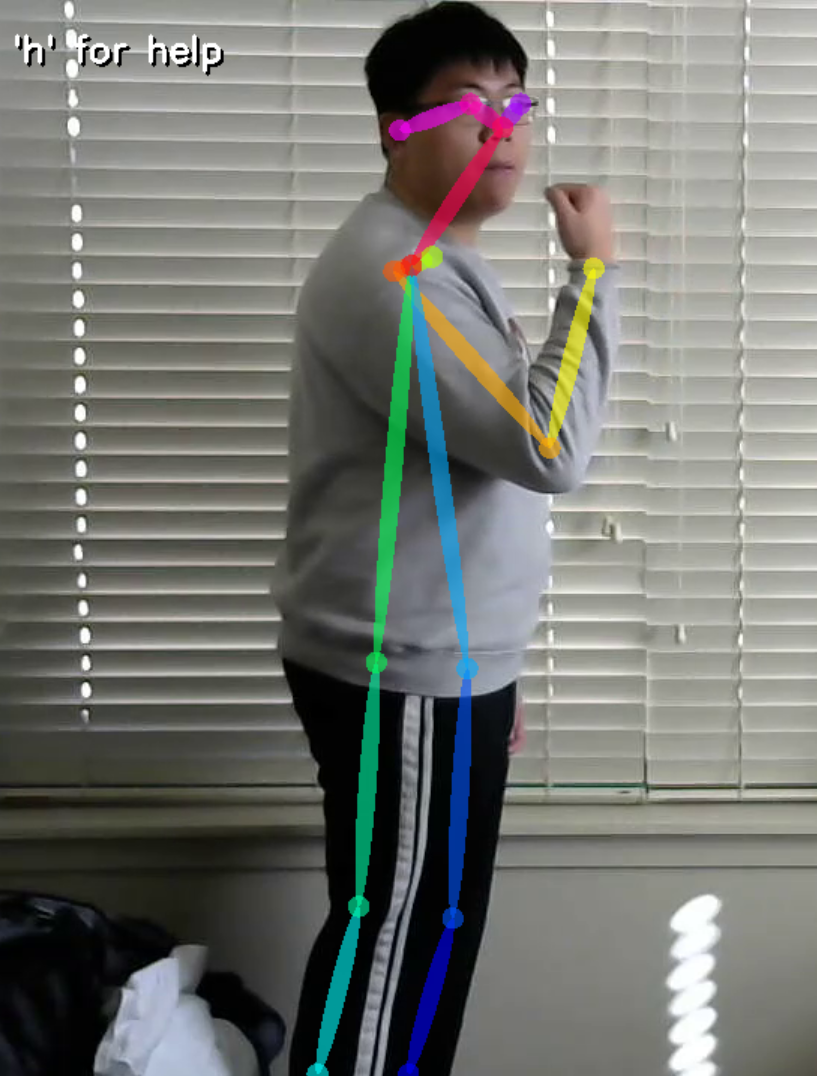}
    
    \includegraphics[width=0.49\linewidth]{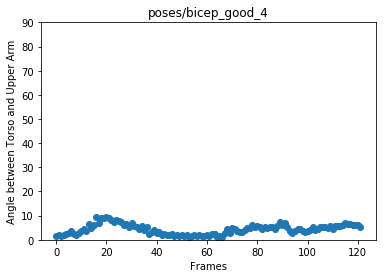}
    \hfill
    \includegraphics[width=0.49\linewidth]{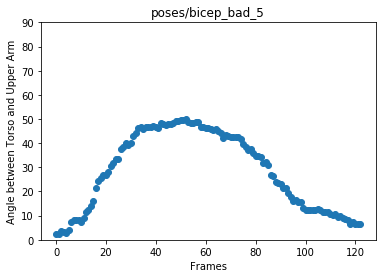}
    \centering
    \caption{A proper bicep curl should focus on the bicep muscle, and only the forearm should move. Left shows proper form, where the upper arm is parallel to the torso: right show improper form, where the upper arm has moved during the curl, commonly due to the strain of too much weight. The bottom plots show the angle between the torso and the upper arm: throughout the exercise, this angle should be small and should not change.}
    \label{fig:bicep_curl_upper_arm}
\end{figure}

Figures \ref{fig:bicep_curl_lift_up} and \ref{fig:bicep_curl_upper_arm} illustrate examples of good and bad form, as well as the pose angle statistics that we use when evaluating a user's exercise.

Below is an example of Pose Trainer's output on a correctly performed bicep curl:

\begin{minted}[breaklines,fontsize=\footnotesize]{text}
python main.py --video videos\bicep_good_1.mp4 --output_folder temp --mode evaluate --exercise bicep_curl
processing video file...
Exercise arm detected as: right.
Upper arm and torso angle range: 21.150955500327434
Upper arm and forearm minimum angle: 40.74447650965106
Exercise performed correctly!
Exercise performed correctly! Weight was lifted fully up, and upper arm did not move significantly.
\end{minted}

Below is an example of Pose Trainer's output on an incorrectly performed bicep curl:

\begin{minted}[breaklines,fontsize=\footnotesize]{text}
C:\Users\stevenzc\Documents\git\pose-trainer>python main.py --video videos\bicep_bad_1.mp4 --output_folder temp --mode evaluate --exercise bicep_curl
processing video file...
Exercise arm detected as: right.
Upper arm and torso angle range: 35.23131076818897
Upper arm and forearm minimum angle: 31.89380019853305
Exercise could be improved:
Your upper arm shows significant rotation around the shoulder when curling. Try holding your upper arm still, parallel to your chest, and concentrate on rotating around your elbow only.
\end{minted}

\subsection{Front Raise}

\begin{figure}[t]
    \centering
    \includegraphics[width=0.4\linewidth]{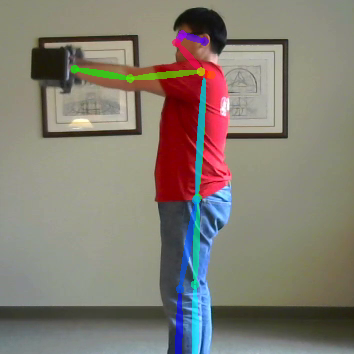}
    \hspace{0.5cm}
    \includegraphics[width=0.4\linewidth]{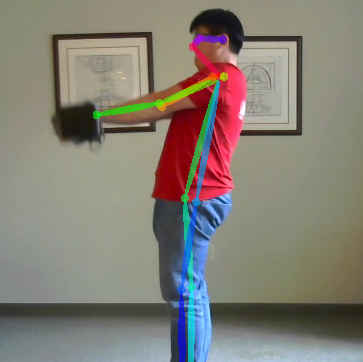}
    
    \includegraphics[width=0.49\linewidth]{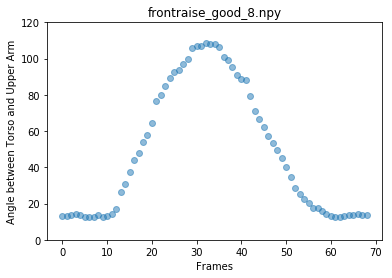}
    \hfill
    \includegraphics[width=0.49\linewidth]{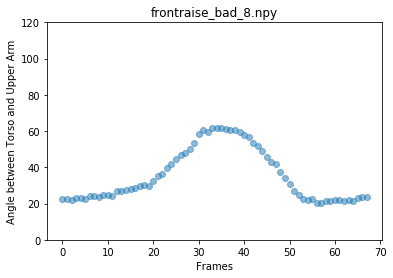}
    \centering
    \caption{Front Raise. In a proper front raise, as shown on the left, the lifter should keep the torso still, and raise the weights to above the shoulders. The corresponding curve shows the wide range of motion as the angle between the arm and the torso. Incorrect form is shown on the right, where the torso moves significantly with the weight and the weights are not lifted high enough.}
    \label{fig:frontraise}
\end{figure}

Dumbbell front raise is a free weight exercise targeting the shoulders, specifically the anterior deltoids. In the front raise, the lifter holds weights at their side, and lifts the weights straight in front of them, with arms mostly straight. The lifter should keep the body still to isolate the exercise to the shoulders, and should lift up to slightly above the shoulders to complete the full range of motion. Common mistakes include straining and using torso movement to help lift, as well as not lifting up to above the shoulders (see Figure \ref{fig:frontraise}).

Our geometric algorithm for front raise measures two things: the horizontal range of motion of the back, and the maximum angle between the torso and the arm. Back motion is measured by measuring the greatest change in vector difference between frames of the exercise, and is used to determine if the user is swinging their upper body to help lift the weight. Torso arm angle is used to determine if the user is lifting all the way up.

For our ML algorithm, we split our 28-video front raise dataset into 16 training examples and 12 test examples. The DTW classifier achieves a perfect F1 score.

Pose trainer will output the following feedback to users for front raise: 

\textit{`Your back shows significant movement. Try keeping your back straight and still when you lift the weight. Consider using lighter weight.'}

\textit{`You are not lifting the weight all the way up. Finish with wrists at or slightly above shoulder level.'}

\subsection{Shoulder Shrug}

\begin{figure}[t]
    \centering
    \includegraphics[width=0.4\linewidth]{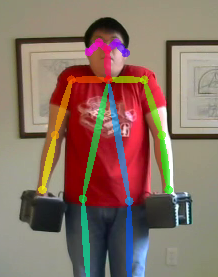}
    \hspace{0.5cm}
    \includegraphics[width=0.38\linewidth]{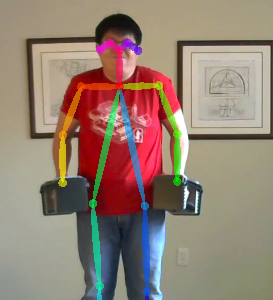}
    
    \includegraphics[width=0.49\linewidth]{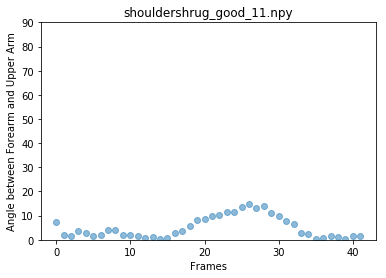}
    \hfill
    \includegraphics[width=0.49\linewidth]{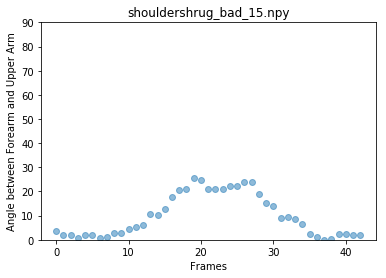}
    \centering
    \caption{Shoulder Shrug. A correct shoulder shrug, as shown on the left, should see significant vertical movement of the shoulder keypoints, while arms should be kept straight. On the right, the lifter is bending their elbows, which is incorrect form since the shoulders are not being isolated. This can be seen with a larger angle between the upper arm and forearm.}
    \label{fig:shouldershrug}
\end{figure}

Dumbbell shoulder shrug is an exercise that focuses on the shoulders, specifically the upper trapezius. The lifter holds dumbbells at their sides and raises their shoulders as much as possible, while keeping the torso rigid and the elbows in a relaxed position. Common mistakes include not going through the full range of motion in the shoulders, and bending the elbows and using the arms when lifting (see Figure \ref{fig:shouldershrug}).

Our geometric algorithm for shoulder shrug measures the range of motion of the shoulders (normalized by torso length), as well as the angle between the upper arm and forearm. If the range of shoulder movement is too low, we flag as the mistake of not going through the full range of motion. If the angle between upper arm and forearm is too small, this means the user is bending too much at the elbow.

For our ML algorithm, we split our 32-video shoulder shrug dataset into 19 training examples and 13 test examples. The DTW classifier achieves an F1 score of 0.85.

Pose trainer will output the following feedback to users for shoulder shrug: 

\textit{`Your shoulders do not go through enough motion. Squeeze and raise your shoulders more through the exercise.'}

\textit{`Your arms are bending when lifting. Keep your arms straight and still, and focus on moving only the shoulders.'}

\subsection{Shoulder Press}

\begin{figure}[t]
    \centering
    \includegraphics[width=0.4\linewidth]{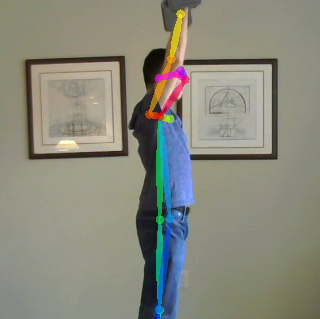}
    \hspace{0.5cm}
    \includegraphics[width=0.4\linewidth]{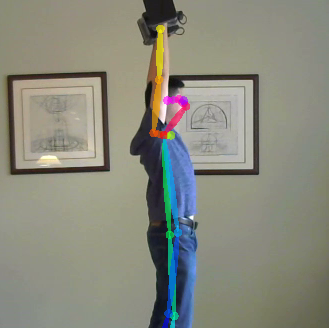}
    
    \includegraphics[width=0.49\linewidth]{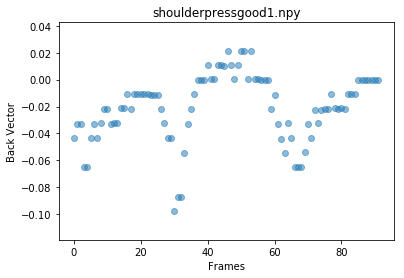}
    \hfill
    \includegraphics[width=0.49\linewidth]{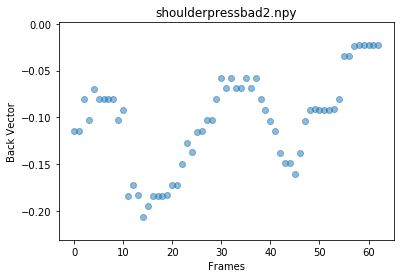}
    \centering
    \caption{Shoulder Press. In a correctly performed shoulder press, as shown on the left, the torso should stay straight and still, and the weights should be slightly in front of the body. Incorrect form includes twisting the torso under the weight, as shown on the right.}
    \label{fig:shoulderpress}
\end{figure}

Dumbbell shoulder press, or overhead press, is a weight training exercise where dumbbells are pressed straight upwards. The exercise works many muscle groups in the shoulders, chest, and upper back. Many mistakes can be made in this exercise: the lifter could raise the weights too far forward or back, move the torso too much when lifting, or not lift the weights high enough (see Figure \ref{fig:shoulderpress}).

Our geometric algorithm for shoulder press measures the range of motion of the back using the neck and hip keypoints, the range of motion of the arm using the elbow and neck keypoints, and the maximum angle achieved between the upper arm and forearm vectors. If the range of motion for the back is too large, we warn the user to keep their back straight. If the location of the elbow is behind their neck, we warn the user to not roll their shoulders during the lift. If the angle between the upper arm and forearm is too small, we warn the user to lift the weight up all the way.

For our ML algorithm, we split our 36-video front raise dataset into 21 training examples and 15 test examples. The DTW classifier achieves an F1 score of 0.73.

Pose trainer will output the following feedback to users for shoulder shrug: 

\textit{`Your back shows significant movement while pressing. Try keeping your back straight and still when you lift the weight.'}

\textit{`You are rolling your shoulders when you lift the weights. Try to steady your shoulders and keep them parallel.'}

\textit{`You are not lifting the weight all the way up. Extend your arms through the full range of motion. Lower the weight if necessary.'}

\begin{table}[t]
\begin{tabular}{rrrrr}
\hline
\multicolumn{1}{l}{} & \textbf{Precision}     & \textbf{Recall}        & \textbf{F1 Score}      & \textbf{Examples}    \\ \hline
\multicolumn{5}{c}{Bicep Curl}                                                     \\ \hline
Correct              & 0.80          & 1.00          & 0.89          & 4           \\
Incorrect            & 1.00          & 0.67          & 0.80          & 3           \\
\textbf{Avg/Total}   & \textbf{0.89} & \textbf{0.86} & \textbf{0.85} & \textbf{7}  \\ \hline
\multicolumn{5}{c}{Front Raise}                                                    \\ \hline
Correct              & 1.00          & 1.00          & 1.00          & 6           \\
Incorrect            & 1.00          & 1.00          & 1.00          & 6           \\
\textbf{Avg/Total}   & \textbf{1.00} & \textbf{1.00} & \textbf{1.00} & \textbf{12} \\ \hline
\multicolumn{5}{c}{Shoulder Shrug}                                                 \\ \hline
Correct              & 1.00          & 0.75          & 0.86          & 8           \\
Incorrect            & 0.71          & 1.00          & 0.83          & 5           \\
\textbf{Avg/Total}   & \textbf{0.89} & \textbf{0.85} & \textbf{0.85} & \textbf{13} \\ \hline
\multicolumn{5}{c}{Shoulder Press}                                                 \\ \hline
Correct              & 0.67          & 0.86          & 0.75          & 7           \\
Incorrect            & 0.83          & 0.62          & 0.71          & 8           \\
\textbf{Avg/Total}   & \textbf{0.76} & \textbf{0.73} & \textbf{0.73} & \textbf{15} \\ \hline
\end{tabular}

\vspace{0.1cm}

\centering
\caption{Confusion matrix for our DTW classification model}
\label{tbl:dtw_comparison}
\end{table}

\section{Conclusion and Future Work}

In this report, we introduce Pose Trainer, an end-to-end computer vision application that uses pose estimation, visual geometry, and machine learning to provide personalized feedback on fitness exercise form. We use the output of pose estimation to evaluate videos of exercises through human pose keypoints. We work with four different exercises, recording training videos for each, and use both geometric heuristic algorithms to provide personalized feedback on specific exercise improvements, as well as machine learning algorithms to automatically determine posture correctness using only labeled input videos. 

We have identified several extensions as strong opportunities for future work past this course project. One path would be to export Pose Trainer to smartphones, building an application that allows users to record a video and get pose feedback at any place or time. Another direction would be to improve the pose feedback, providing specific suggestions on where the user's pose needs improvement (e.g., back, neck, shoulders), and suggesting targeted action. Finally, we could work on improved graphics, for instance, showing the user their labeled pose diagram, and comparing to the labeled pose diagram of a ground truth trainer.

{\small
\bibliographystyle{ieee}
\bibliography{egbib}
}

\end{document}